\documentclass[10pt,twocolumn,letterpaper]{article}

\PassOptionsToPackage{unicode}{hyperref}
\PassOptionsToPackage{naturalnames}{hyperref}

\usepackage{iccv}
\usepackage[utf8]{inputenc}
\usepackage{times}
\usepackage{graphicx}
\usepackage{subcaption}
\usepackage{color}
\usepackage{amsmath}
\usepackage{amssymb}
\usepackage{color,soul}
\usepackage{algorithm}
\usepackage{algorithmic}
\usepackage{multirow}

\usepackage[pagebackref=true,breaklinks=true,colorlinks,bookmarks=false]{hyperref}

\usepackage{xstring}
\usepackage{fdsymbol}
\usepackage{bbding}

\newcommand{\SymbolCircle}[1]{ $\color{#1} \medcircle $ }
\newcommand{\SymbolSquare}[1]{ $\color{#1} \medsquare $ }
\newcommand{\SymbolDiamond}[1]{ $\color{#1} \meddiamond $ }
\newcommand{\SymbolCross}[1]{ $\color{#1} \times $ }
\newcommand{\SymbolPlus}[1]{ $\color{#1} \plus $ }
\newcommand{\SymbolAsterisk}[1]{ {\tiny\color{#1} \Asterisk } }
\newcommand{\SymbolTriangleU}[1]{ $\color{#1} \triangle$ }
\newcommand{\SymbolTriangleL}[1]{ $\color{#1} \triangleleft$ }
\newcommand{\SymbolTriangleR}[1]{ $\color{#1} \triangleright$ }
\newcommand{\SymbolTriangleD}[1]{ $\color{#1} \triangledown$ }
\newcommand{\SymbolPentagram}[1]{ $\color{#1} \medwhitestar$ }
\newcommand{\SymbolHexagram}[1]{ {\tiny\color{#1} \DavidStar} }

\definecolor{TrackerSymbolColor1}{rgb}{0.00,0.57,1.00}
\definecolor{TrackerSymbolColor2}{rgb}{0.00,1.00,0.57}
\definecolor{TrackerSymbolColor3}{rgb}{0.29,0.00,1.00}
\definecolor{TrackerSymbolColor4}{rgb}{0.29,1.00,0.00}
\definecolor{TrackerSymbolColor5}{rgb}{1.00,0.00,0.00}
\definecolor{TrackerSymbolColor6}{rgb}{1.00,0.00,0.86}
\definecolor{TrackerSymbolColor7}{rgb}{1.00,0.86,0.00}
\newcommand{\TrackerSymbol}[1]{%
\IfStrEqCase{#1}{
{KCF}{ \SymbolCircle{TrackerSymbolColor5}  }
{ASMS}{ \SymbolCross{TrackerSymbolColor7}  }
{Struck}{ \SymbolAsterisk{TrackerSymbolColor4}  }
{MIL}{ \SymbolTriangleD{TrackerSymbolColor2}  }
{CMT}{ \SymbolDiamond{TrackerSymbolColor1}  }
{FoT}{ \SymbolPlus{TrackerSymbolColor3}  }
{DSST}{ \SymbolTriangleL{TrackerSymbolColor6}  }
{FragTrack}{ \SymbolPentagram{TrackerSymbolColor5}  }
{CT}{ \SymbolTriangleR{TrackerSymbolColor7}  }
{L1APG}{ \SymbolSquare{TrackerSymbolColor4}  }
{Staple}{ \SymbolTriangleU{TrackerSymbolColor2}  }
{SiameseFC}{ \SymbolHexagram{TrackerSymbolColor1}  }
{LGT}{ \SymbolCircle{TrackerSymbolColor3}  }
{IVT}{ \SymbolCross{TrackerSymbolColor6}  }
{MEEM}{ \SymbolAsterisk{TrackerSymbolColor5}  }
{ASLA}{ \SymbolTriangleD{TrackerSymbolColor7}  }
{SRDCF}{ \SymbolDiamond{TrackerSymbolColor4}  }
}
}

\graphicspath{{./figures/}}
\newcommand{\first}[1]{%
    \bf{\color{red}#1}%
}
\newcommand{\second}[1]{%
    \em{\color{blue}#1}%
}

\newcommand{\scpos}[1]{%
    \bf{\color{green}#1}%
}

\newcommand{\scnegs}[1]{%
    \em{\textcolor[rgb]{1,0.5,0}{#1}}%
}
\newcommand{\scnegm}[1]{%
    \em{\textcolor[rgb]{1,0.1,0}{#1}}%
}
\newcommand{\scnegl}[1]{%
    \em{\textcolor[rgb]{0.7,0,0}{\bf #1}}%
}

\newcommand{\cell}[2][c]{%
  \begin{tabular}[#1]{@{}c@{}}#2\end{tabular}}

\iccvfinalcopy

\def\iccvPaperID{1352}

\ificcvfinal\fi
\begin{document}

\title{Beyond standard benchmarks: Parameterizing performance evaluation \\ in visual object tracking}

\author{Luka \v{C}ehovin Zajc, Alan Luke\v{z}i\v{c}, Ale\v{s} Leonardis, Matej Kristan\\
University of Ljubljana\\
Faculty of computer and information science\\
{\tt\small $\{$luka.cehovin,alan.lukezic,ales.leonardis, matej.kristan$\}$@fri.uni-lj.si}
}

\maketitle


\begin{abstract}
Object-to-camera motion produces a variety of apparent motion patterns that significantly affect performance of short-term visual trackers. Despite being crucial for designing robust trackers, their influence is poorly explored in standard benchmarks due to weakly defined, biased and overlapping attribute annotations. In this paper we propose to go beyond pre-recorded benchmarks with post-hoc annotations by presenting an approach that utilizes omnidirectional videos to generate realistic, consistently annotated, short-term tracking scenarios with exactly parameterized motion patterns. We have created an evaluation system, constructed a fully annotated dataset of omnidirectional videos and the generators for typical motion patterns. We provide an in-depth analysis of major tracking paradigms which is complementary to the standard benchmarks and confirms the expressiveness of our evaluation approach. 
\end{abstract}

\section{Introduction}  \label{sec:introduction} 

Single-target visual object tracking has made significant progress in the last decade. To a large extent this can be attributed to the adoption of benchmarks, through which common evaluation protocols, datasets and baseline algorithms have been established. Starting with PETS initiative~\cite{Young2005}, several benchmarks on general single-target short-term tracking have been developed since, most notably OTB50~\cite{otb_cvpr2010}, VOT2013~\cite{kristan_vot2013}, ALOV300+~\cite{alov_pami2014}, VOT2014~\cite{kristan_vot2014, kristan_vot_tpami2016}, VOT2015~\cite{kristan_vot2015} OTB100~\cite{otb_pami2015}, TC128~\cite{templecolor_tip2015} and VOT2016~\cite{kristan_vot2016}. 

\begin{figure}[!t]
\centering
\includegraphics[width=0.8\linewidth]{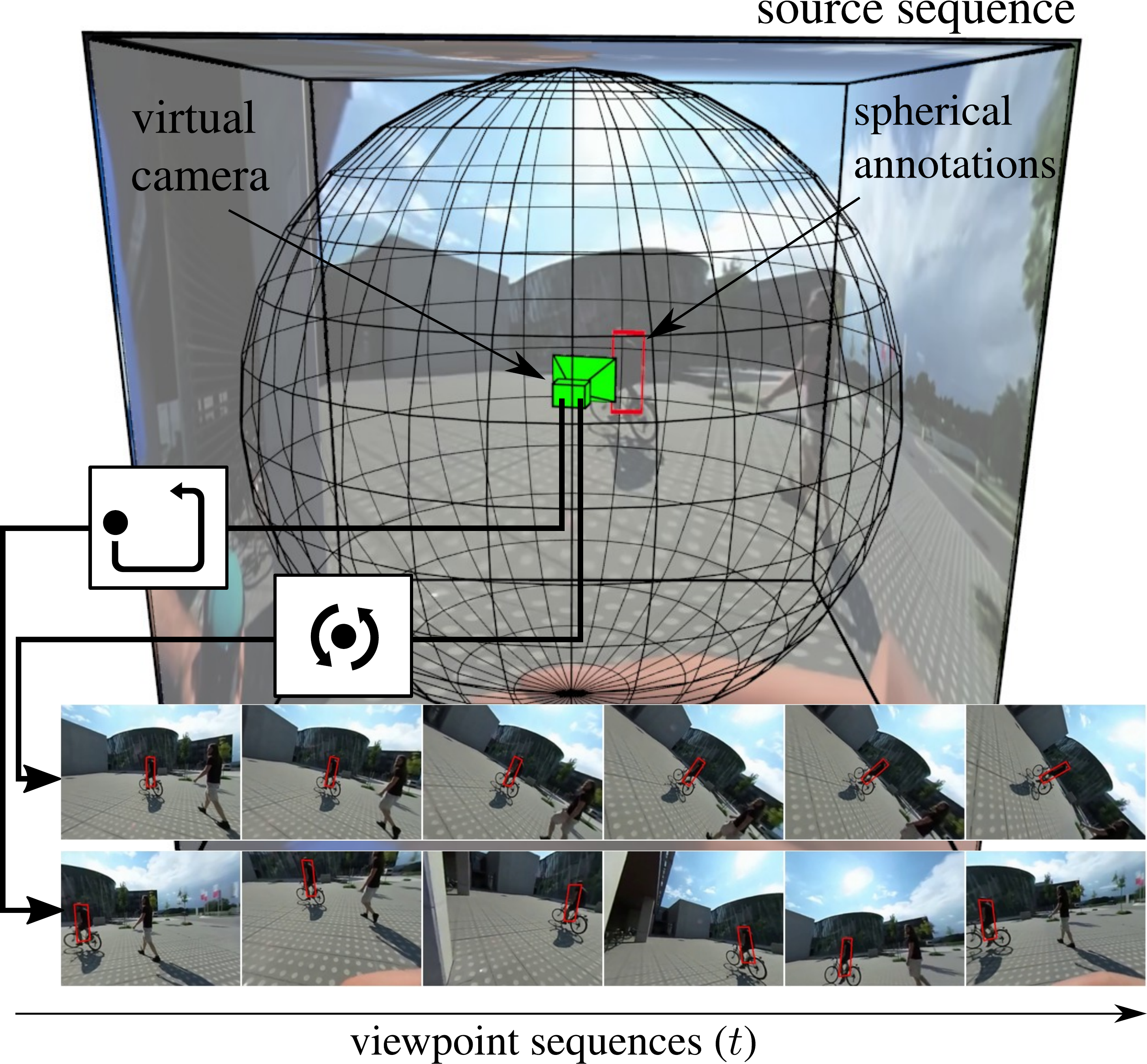}
\caption{By re-parameterizing camera trajectory, a single 360° video produces various 2D viewpoint sequences  with unique apparent motion patterns.
\label{fig:concept}}
\end{figure} 

The recent benchmarks~\cite{kristan_vot2016, uav_benchmark_eccv2016} report that, apart from the obvious situations like full occlusions, the trackers' performance is largely affected by the {\em apparent motion}, i.e., object motion with respect to the camera. The complexity of apparent motion patterns varies in realistic applications. An automated video-conferencing system largely observes translational motions, a drone circling over a target induces a large off-center rotational pattern, while water movement induces periodic scale changes in underwater robotic vessels. In some trackers, the translational motions are addressed by motion models. But compositions of scale changes, rotations and off-center translations are often assumed to be addressed by the visual models and localization techniques. This aspect is left largely unexplored in standard benchmarks, which are dominated by a handful of motion patterns and cannot fully expose the limitations of existing trackers. Advances in short-term tracking therefore call for accurate parametrization of apparent motion patterns in test sequences.

 



Dataset variation, systematic organization and low redundancy are crucial for practical evaluation, as argued by the recent work on test-data validation in computer vision~\cite{Zendel_2015_ICCV}. Established benchmarks in visual tracking approach this requirement by increasing the number of sequences~\cite{alov_pami2014}, applying advanced dataset construction methodologies~\cite{kristan_vot2016} and by annotating entire sequences or even individual frames with visual attributes~\cite{otb_pami2015,kristan_vot_tpami2016}. While such {\em bottom-up} approach is suitable for determining overall ranking of algorithms it is insufficient to study the performance of modern trackers along different motion patterns. Benchmarks contain annotated frames (or entire sequences) with only few attributes that correspond to motion patterns, which are only binary, non-parameterized and subject to human annotator bias. Additionally, accurate attribute-wise analysis is difficult due to the attribute cross-talk, meaning that multiple attributes occur at the same interval in a sequence (e.g., object rotation and rapid translation), which prohibits establishing a clear causal relation between a single motion pattern and tracker design performance. In principle, computer graphics generated sequences~\cite{vig_cvpr2016, uav_benchmark_eccv2016} offer full camera control, however the level of realism in object motion and appearance in such sequences still presents a limitation for performance evaluation of general tracking methods.

Our work addresses the 
limitations of the traditional benchmarks by proposing a framework for {\em top-down} construction of test sequences through parametrization of apparent motion patterns. A virtual camera model that utilizes omnidirectional videos is introduced to generate photo-realistic, consistently annotated short-term tracking scenarios (Figure~\ref{fig:concept}). The exact specification of parameterized motion patterns guarantees a clear causal relation between the generated apparent motion and the tracking performance change. This enables fine-grained performance analysis and can be used complementary to the existing benchmarks to offer an 
in-depth analysis of tracking approaches. 
 
\textbf{Contributions.} Our contributions are three-fold. (1) We propose a new performance evaluation paradigm based on generation of realistic sequences with high degree of motion pattern parametrization from annotated omnidirectional videos. (2) We have constructed a new apparent motion benchmark for short-term single-target trackers. A new dataset with per-frame target annotation in omnidirectional videos adding up to 17537 frames and generators of twelve motion patterns are introduced. (3) We have evaluated 17 state-of-the-art trackers from recent benchmarks categorized in major tracking paradigms~\cite{otb_pami2015, kristan_vot2016} and provide insights not available in standard benchmarks. The new benchmark, the results and the corresponding software will be made publicly available and are expected to significantly affect future developments in single-target tracking.

\textbf{Structure.} The paper is organized as follows, in Section~\ref{sec:related} we review related work, in Section~\ref{sec:methodology} we describe our sequence parameterization framework, in Section~\ref{sec:motben} we present the proposed benchmark, in Section~\ref{sec:results} we present experimental results, and in Section~\ref{sec:conclution} we discuss our findings and make concluding remarks.

\section{Related work}
\label{sec:related}

Modern short-term tracking benchmarks ~\cite{otb_pami2015, kristan_vot2013, kristan_vot_tpami2016, kristan_vot2015, kristan_vot2016, alov_pami2014, uav_benchmark_eccv2016} acknowledge the importance of motion related attributes and support evaluation with respect to these. However, the evaluation capability significantly depends on the attribute presence and distribution, which is often related to the sequence acquisition. In application-oriented benchmarks like~\cite{uav_benchmark_eccv2016} the attribute distribution is necessarily skewed by the application domain. Some general benchmarks~\cite{otb_pami2015, alov_pami2014} thus include a large number of sequences from various domains. But since the sequences are post-hoc annotated, the dataset diversity is hard to achieve. A recent benchmark~\cite{kristan_vot2016} addressed this by considering the attributes already at the sequence collection stage and applied an elaborate methodology for automatic dataset construction.

The strength of per-attribute evaluation depends on the annotation approach. In most benchmarks~\cite{otb_pami2015, alov_pami2014, uav_benchmark_eccv2016} all frames are annotated by an attribute even if it occupies only a part of the sequence. Kristan et al.~\cite{kristan_vot_tpami2016} argued that this biases per-attribute performance towards average performance and proposed a per-frame annotation to reduce the bias. However, a single frame might still contain several attributes, resulting in the attribute cross-talk bias. 

The use of computer graphics in training and evaluation has recently been popularized in computer vision. Mueller et al.~\cite{uav_benchmark_eccv2016} propose online virtual worlds creation for drone tracking evaluation, but using only a single type of the object, without motion parametrization, produces a low level of realism. Vig et al.~\cite{vig_cvpr2016} address the virtual worlds realism levels, ambient parametrization learning and performance evaluation, however, only for vehicle detection.

Our work is positioned between the standard benchmarking approaches and the synthetic sequence generation. By using {\em real imagery} we retain photo-realism of standard benchmarks. Our approach simultaneously enables parameterization of apparent motion thus opening new possibilities for in-depth evaluation that is complementary to existing benchmarks.   

\section{Sequence parametrization}
\label{sec:methodology}

Two key concepts are introduced by our apparent-motion evaluation methodology: a {\em source sequence} and a {\em viewpoint sequence}. A source sequence is an omnidirectional video that simultaneously captures 360 degree field of view. The video is stored as a projection onto a spectator-centered sphere, i.e., $\mathcal{S} = \{ S_t \}_{t = 1:N}$, where $S_t$ is a projection at frame $t$. Such representation allows to generate arbitrary views of a 3D scene from the point of observation. 

A viewpoint sequence is a sequence of images obtained from a spherical representation by projection into a pinhole camera, i.e., $\mathcal{I} = \{ I_t \}_{t = 1:N}$. The camera model has adjustable rotation and focal length parameters, thereby defining the state of the camera at time $t$ as
\begin{equation}
	C_t = [\alpha_t, \beta_t, \gamma_t, f_t],
\end{equation}
where the first three parameters are the Euler angles and $f_t$ denotes the focal length. Each frame in a viewpoint sequence is therefore the result of the corresponding image in the source sequence and the camera parameters, i.e. $I_t = p_\mathrm{cam}(S_t;C_t)$.
 
The ground truth object state in each frame is specified in a viewpoint-agnostic spherical coordinate system, i.e., $\mathcal{A} = \{ A_t \}_{t = 1:N}$. Following the VOT Challenge protocol~\cite{kristan_vot_tpami2016} the state is defined as a rectangle using four-points $A_t = \{ \theta^i_t, \rho^i_t \}_{i = 1:4}$. Given a pinhole camera viewpoint parameters $C_t$, the ground truth $A_t$ is projected into the image plane by projective geometry, i.e., $G_t = p_\mathrm{gt}( A_t; C_t )$.

The camera parameters $C_t$ are continually adjusted during the creation of the viewpoint sequence to keep the projected object within the field of view, thus satisfying the short-term tracking constraint. The camera viewpoint is adjusted via a {\em camera controller} $p_\mathrm{con}(\cdot, \cdot)$ that applies a prescribed motion pattern $E$ and maps the object ground truth state into camera parameters while satisfying the short-term tracking constraint, i.e.,
\begin{equation}
	p_\mathrm{con}(A_t, E, t) \mapsto C_t.
\end{equation}
Depending on the pattern type specification, the controller continually adjusts camera-to-object position and generates various apparent object motions.

\subsection{Evaluation framework}\label{sec:framework}

The evaluation framework implements the \textit{VOT supervised evaluation mode}~\cite{Cehovin_TIP2016} and the VOT~\cite{kristan_vot_tpami2016,kristan_vot2016} performance evaluation protocol, which allows full use of long sequences. In this evaluation mode, a tracker is initialized and re-set upon drifting off the target. Stochastic trackers are run multiple times and the results are averaged. 

\begin{figure}[ht]
\centering
\includegraphics[width=\linewidth]{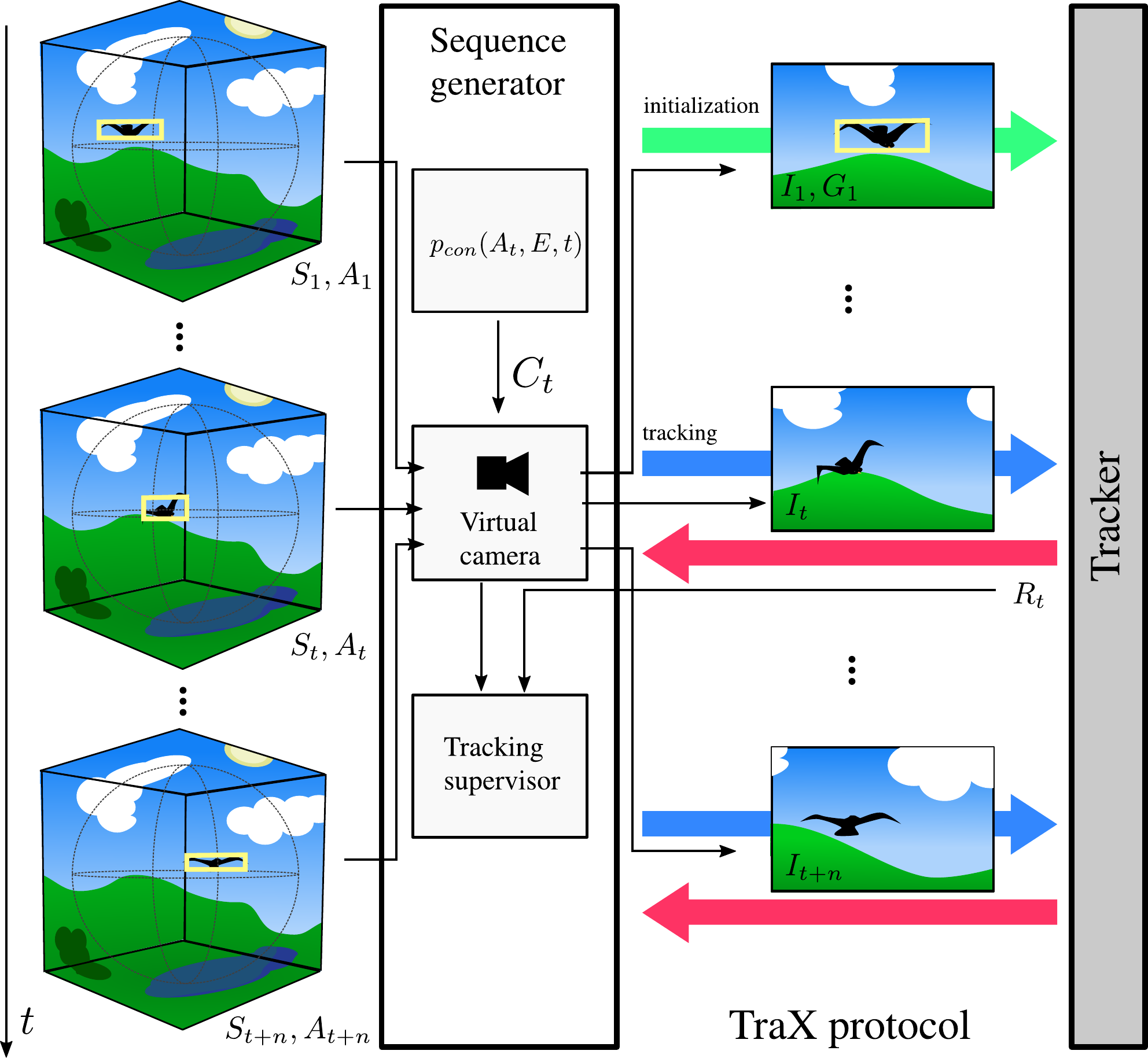}\\
\caption{The evaluation framework. The sequence generator constructs a viewpoint sequence with corresponding ground truths according to the motion pattern. Tracker reports predicted region in each frame to the evaluator for automatic failure detection.}
\label{fig:system}
\end{figure} 

The following functionality is required by the supervised experiment mode: (1) reproducible sequence generation and (2) bi-directional tracker-evaluator communication. The viewpoint sequence and the 2D ground truth are therefore generated on the fly during the evaluation and are reproducible for each time-step. The communication between the the evaluator and the tracker is implemented through the state-of-the-art TraX~\cite{cehovin_trax} communication protocol. Our evaluation framework is summarized in Figure~\ref{fig:system}.
 
\section{Apparent-motion patterns benchmark}\label{sec:motben}

Our motion parametrization framework is demonstrated on a novel single-target visual object tracking benchmark for isolated apparent-motion patterns (AMP). The benchmark contains very long 15 omnidirectional sequences (adding up to 17537 frames) and specifies twelve motion types.
 
\subsection{Dataset acquisition}

\begin{figure}[h]
\centering
\includegraphics[width=0.95\linewidth]{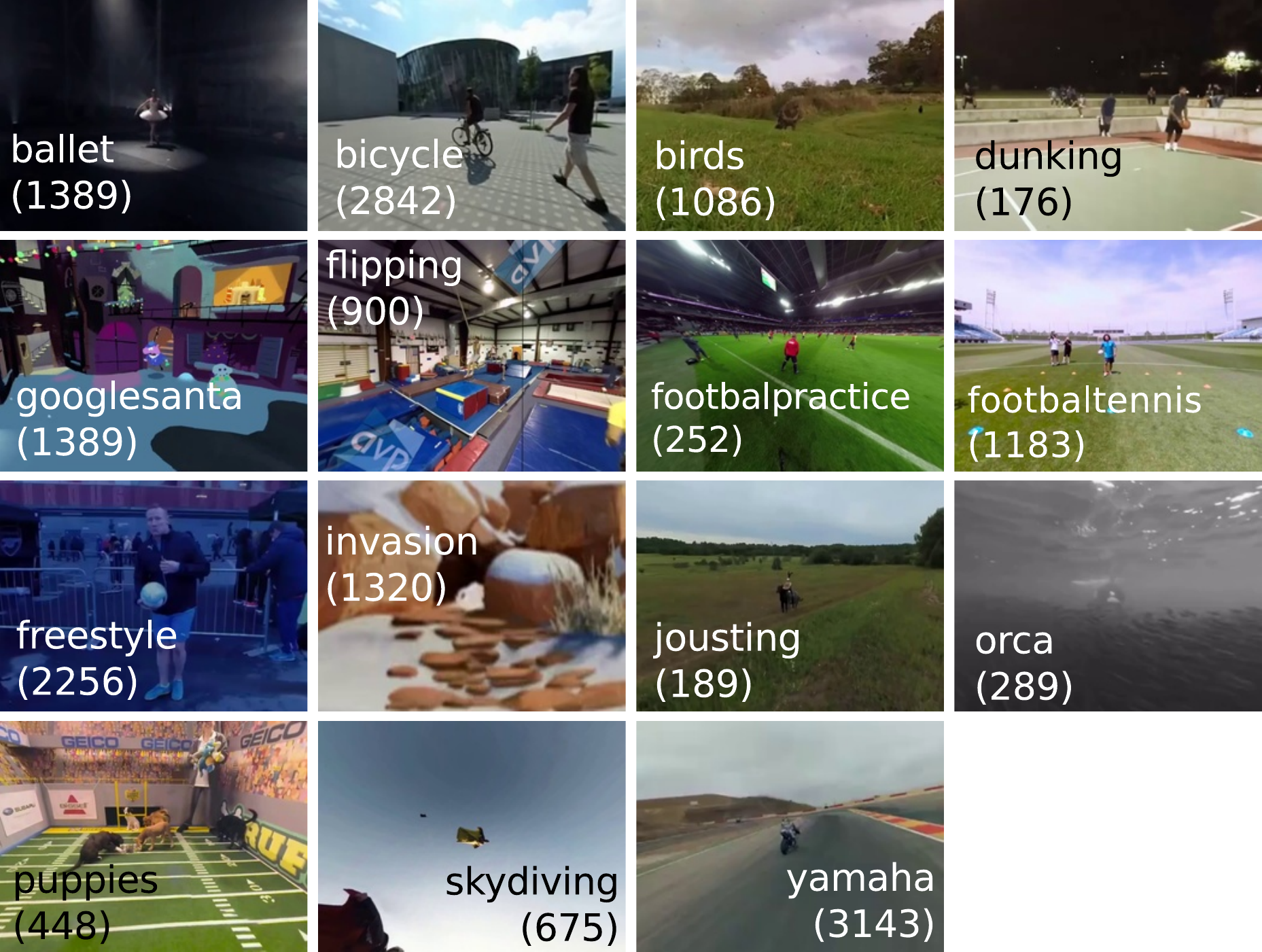}\\
\caption{A preview of 360 degree sequences in the dataset from the view that centers the target. 
}
\label{fig:dataset}
\end{figure}

The new dataset contains fifteen omnidirectional videos with an average video length of 1169 frames, amounting to 17537 frames. The videos were mostly selected from a large collection of 360 degree videos available on YouTube. To maximize the target diversity, we recorded additional sequences using Ricoh Theta 360 degree camera. Videos were converted to a cube-map projection and encoded with MP4 H.264 codec. Each frame of the video was manually annotated by a rectangular region encoded in spherical coordinates using an annotation tool specifically designed for this use case. Some of the viewpoint frame examples of individual source sequences are shown in Figure~\ref{fig:dataset}.

\subsection{Motion patterns specification}\label{sec:motion_types}

We consider six motion pattern classes that reflect typical dynamic relations between an object (target) and a camera:

\begin{figure}[h]
\centering
\includegraphics[width=0.9\linewidth]{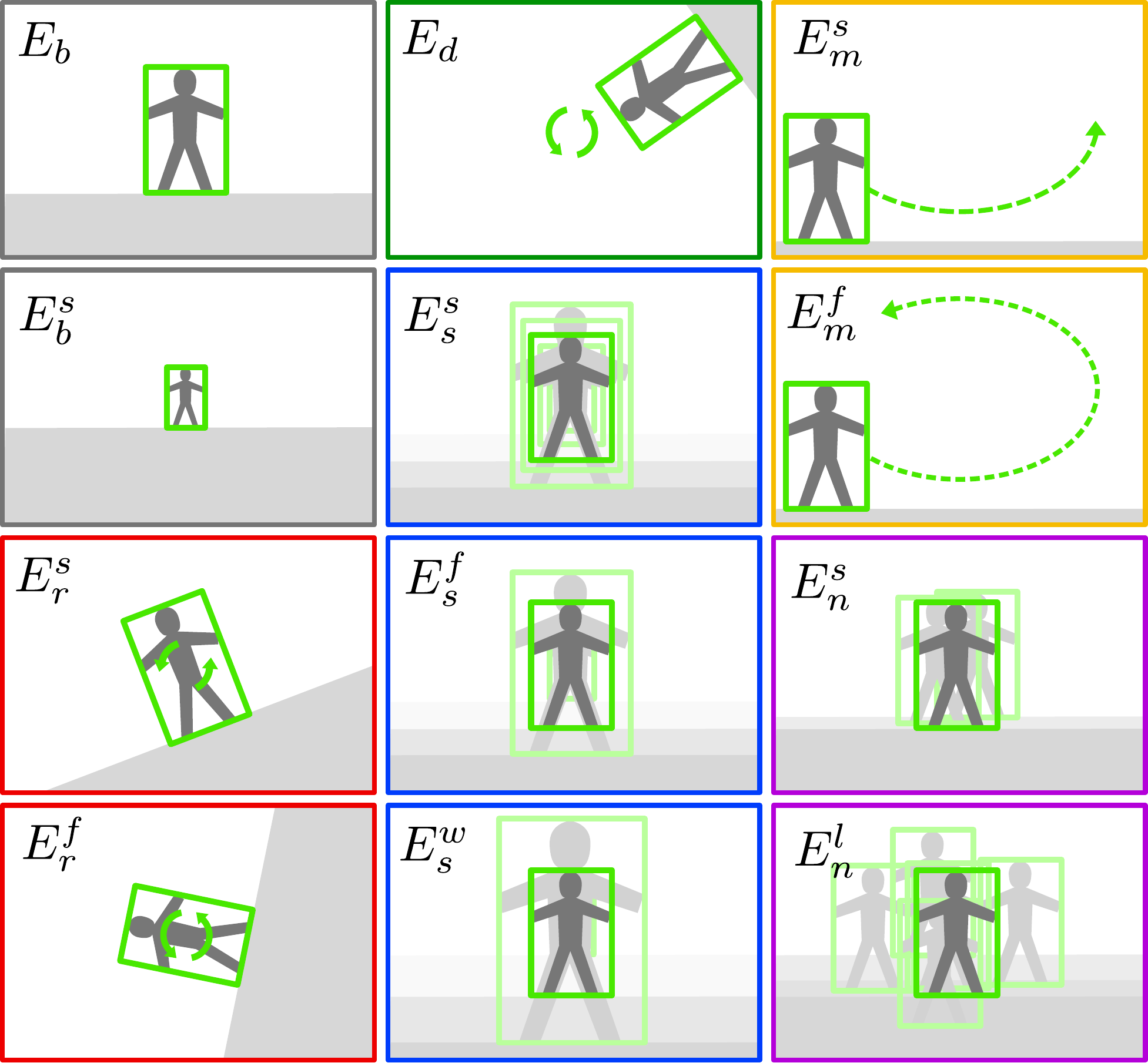}\\
\caption{The twelve apparent-motion patterns in the AMP benchmark.}\label{fig:motion_classes}
\end{figure} 

\noindent \textbf{Stabilized setup}, denoted as $\mathrm{E}_\mathrm{b}$, keeps the object position at image center and adjusts the camera distance to keep the object diagonal constant at $70$ pixels. A variant with a diagonal constant at $35$ pixels is considered as well to test tracking objects from far away, $\mathrm{E}_\mathrm{b}^{s}$.

\noindent \textbf{Centered rotation setup}, denoted as $\mathrm{E}_\mathrm{r}$, fixes the object center and the scale as $\mathrm{E}_\mathrm{b}$ and then rotates the camera around the optical axis. Two variants, with low and high rotation speeds, $\mathrm{E}_\mathrm{r}^\mathrm{s}$ and $\mathrm{E}_\mathrm{r}^\mathrm{f}$, respectively, are considered.

\noindent \textbf{Displaced rotation setup}, denoted as $\mathrm{E}_\mathrm{d}$, displaces the object center and then rotates the camera around its optical axis.

\noindent \textbf{Scale change setup}, denoted as $\mathrm{E}_\mathrm{s}$, fixes the center and then periodically changes the scale by a cosine function with amplitude oscillation around the nominal scale of $\mathrm{E}_\mathrm{b}$. Two variants, i.e., with a low, $\mathrm{E}_\mathrm{s}^\mathrm{s}$, and a high frequency, $\mathrm{E}_\mathrm{s}^\mathrm{f}$, but equally moderate amplitude are considered. Another variant with a moderate frequency but large amplitude, $\mathrm{E}_\mathrm{s}^\mathrm{w}$, is considered as well.

\noindent \textbf{Planar motion setup}, denoted as $\mathrm{E}_\mathrm{m}$, displaces the camera from the object center and performs circular motion in image plane. A variant with low -- $\mathrm{E}_\mathrm{m}^\mathrm{s}$ and high -- $\mathrm{E}_\mathrm{m}^\mathrm{f}$ frequency are considered.

\noindent \textbf{Translation noise setup}, denoted as $\mathrm{E}_\mathrm{n}$, fixes the center and the scale as in $\mathrm{E}_\mathrm{b}$ then randomly displaces the center by drawing a displacement vector from a normal distribution. Two variants, one with small, $\mathrm{E}_\mathrm{n}^\mathrm{s}$, and one with large, $\mathrm{E}_\mathrm{n}^\mathrm{l}$, noise are considered.

The variations of six motion classes result in 12 different motion patterns, which are illustrated in Figure~\ref{fig:motion_classes}. While these patterns may seem synthetic, they actually occur in many active-camera robotics scenarios, e.g. a drone circling over an observed target (rotation) or an autonomous boat being swayed by the sea (scale change).
Note that each omnidirectional video in our dataset creates a sequence with specific motion parameters. Thus the effect of each motion pattern is evaluated on all frames without being influenced by presence of other patterns, establishing clear casual relationships between the patterns and the tracker's performances.

\subsection{Benchmark comparison}

A comparison of our proposed AMP with most popular standard benchmarks is summarized in Table~\ref{tab:tracking_datasets_overview}. The values under $\mathrm{MAC}$ indicate the percentage of frames in the dataset with at least a single motion attribute. The motion attributes are most frequent in the AMP ($100 \%$ coverage) and UAV123~\cite{uav_benchmark_eccv2016} ($96 \%$ coverage). To reflect the dataset size in motion evaluation, we compute the number of effective frames per attribute (FPA). This measure counts the number of frames that contain a particular motion attribute, where each frame contributes with weight inversely proportional to the number of motion attributes it contains. The FPA is highest for an application-specific UAV123~\cite{uav_benchmark_eccv2016} (27107). Among the general tracking benchmarks, this value is highest for the proposed AMP (17537), which exceeds the second largest (OTB100~\cite{otb_pami2015}) by over $30\%$.

\begin{figure*}[ht]
\begin{subfigure}{0.89\textwidth}
\includegraphics[width=\linewidth]{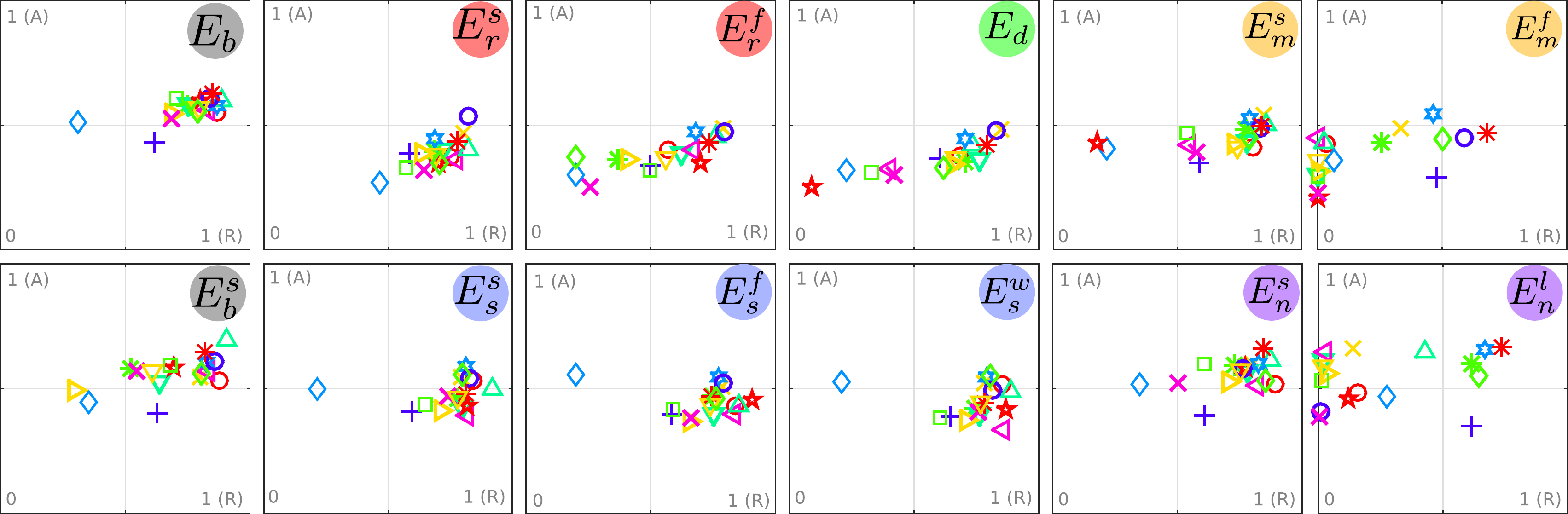}
\end{subfigure}%
\begin{subfigure}{0.10\textwidth}
{\scriptsize \begin{tabular}{l}
\TrackerSymbol{ASLA} ASLA\\
\TrackerSymbol{ASMS} ASMS\\
\TrackerSymbol{CMT} CMT\\
\TrackerSymbol{CT} CT\\
\TrackerSymbol{DSST} DSST\\
\TrackerSymbol{FoT} FoT\\
\TrackerSymbol{FragTrack} FragTrack\\
\TrackerSymbol{IVT} IVT\\
\TrackerSymbol{KCF} KCF\\
\TrackerSymbol{L1APG} L1APG\\
\TrackerSymbol{LGT} LGT\\
\TrackerSymbol{MEEM} MEEM\\
\TrackerSymbol{MIL} MIL\\
\TrackerSymbol{SRDCF} SRDCF\\
\TrackerSymbol{SiameseFC} SiamFC\\
\TrackerSymbol{Staple} Staple\\
\TrackerSymbol{Struck} Struck\\
\end{tabular} }
\end{subfigure}
\caption{A-R plots for each experiment in the benchmark. The vertical axis denotes accuracy and the horizontal axis denotes robustness. The {\em sensitivity} visualization parameter was set to $100$ frames in all plots.}\label{fig:ar_plots}
\end{figure*} 

The FPA alone does not fully reflect the evaluation strength since it does not account for the attribute cross-talk. A lower bound on the cross-talk is reflected by the $\mathrm{INTER}$ measure that shows a percentage of motion-annotated frames with at least two motion attributes. The measure shows that well over half of the frames in UAV123~\cite{uav_benchmark_eccv2016} ($78\%$) and OTB100~\cite{otb_pami2015} ($63\%$) suffer from the attribute cross-talk. The cross-talk is lowest for the proposed AMP ($0\%$), ALOV~\cite{alov_pami2014} ($0\%$) and VOT2016~\cite{kristan_vot2016} ($32\%$). 

Most existing benchmarks are annotated by four motion pattern types. The proposed AMP benchmark contains approximately three times more motion pattern types than existing benchmarks. The existing benchmarks lack motion pattern quantification (e.g., the extent of \textit{speed} in attribute fast motion), which results in inconsistent definitions across benchmarks. In contrast, the motion patterns are objectively defined through their parametrization in the proposed AMP benchmark. 
 
\begin{table}[htb]
\centering
\footnotesize
\caption{Comparison of AMP with popular recent tracking benchmarks: ALOV300+~\cite{alov_pami2014}, OTB100~\cite{otb_pami2015}, UAV123~\cite{uav_benchmark_eccv2016} and VOT2016~\cite{kristan_vot2016}. Best, second best and third best values are shown in red, blue and green, respectively.}
\label{tab:tracking_datasets_overview}
\begin{tabular}{lccccc}
Dataset & \cite{alov_pami2014} & \cite{otb_pami2015} & \cite{uav_benchmark_eccv2016} & \cite{kristan_vot2016} & AMP \\
\hline
\small{MAC (\%)} & 19 & 88 & \second{96} & 61 & \first{100} \\
\small{FPA} & 4275 & 12929 & \first{27107} & 4366 & \second{17537} \\
\small{INTER (\%)} & \first{0} & 63 & 78 & \second{32} & \first{0} \\
\small{Motion classes} & {3} & {3} & {3} & {3} & \first{6} \\
\small{Motion patterns} & \second{4} &  \second{4} & \second{4} & {3} & \first{12} \\
\small{Parameterized} & no & no & no & no & \first{yes} \\
\small{Per-frame} & no & no & no & \first{yes} & \first{yes}
\end{tabular}
\end{table} 

\subsection{Performance measures}

The tracking performance is measured by the VOT~\cite{kristan_vot2015} measures: tracker accuracy (A), robustness (R) as well as the expected average overlap (EAO). The accuracy measures the overlap between the output of the tracker and the ground truth bounding box during periods of successful tracking, while the robustness measures the number of times a tracker failed and required re-initialization~\cite{Cehovin_TIP2016}. The expected average overlap score is an estimator of the average overlap on a typical short-term sequence a tracker would obtain without reset~\cite{kristan_vot2015}. All scores are calculated on per-sequence basis and averaged with weights proportional to the sequence length.

\section{Evaluation and results}
\label{sec:results}

To demonstrate the verbosity of the AMP benchmarks we have evaluated 17 trackers. Each trackers was evaluated on a total of $210444$ frames, which makes this the largest fine-grained motion-related tracker evaluation to date.

\subsection{Trackers tested}

A set of 17 trackers was constructed by considering baseline and top-performing representatives on recent benchmarks~\cite{otb_pami2015, kristan_vot2016} from the following 6 broad classes of trackers. \textit{(1) Baselines} include standard discriminative and generative trackers MIL~\cite{babenko_mil}, CT~\cite{zhang_ct}, IVT~\cite{ross_ivt}, and FragTrac~\cite{adam2006fragtrack}, a state-of-the-art mean-shift tracker ASMS~\cite{vojir_asms}, and Struck SVM tracker~\cite{hare_struck}. \textit{(2) Correlation filters} include the standard KCF~\cite{henriques2015tracking} and three top-performing correlation filters on VOT2016~\cite{kristan_vot2016} -- DSST~\cite{danelljan2014accurate}, Staple~\cite{staple_cvpr2016} and SRDCF~\cite{danelljan2015srdcf}. \textit{(3) Sparse trackers} include top-performing sparse trackers L1APG~\cite{l1apg_cvpr2012} and ASLA~\cite{asla_cvpr2012}. \textit{(4) Part-based trackers} include the recent state-of-the-art CMT~\cite{cmt_cvpr2015}, LGT~\cite{lgt_tpami2013}, FoT~\cite{vojir_fot}. In addition the set comprises a state-of-the-art \textit{(5) Hybrid tracker} MEEM~\cite{meem_eccv2014} and \textit{(6) ConvNet tracker} SiamCF~\cite{siamese}.


\subsection{Experimental results}

\begin{figure}[ht]
\centering
\includegraphics[width=\linewidth]{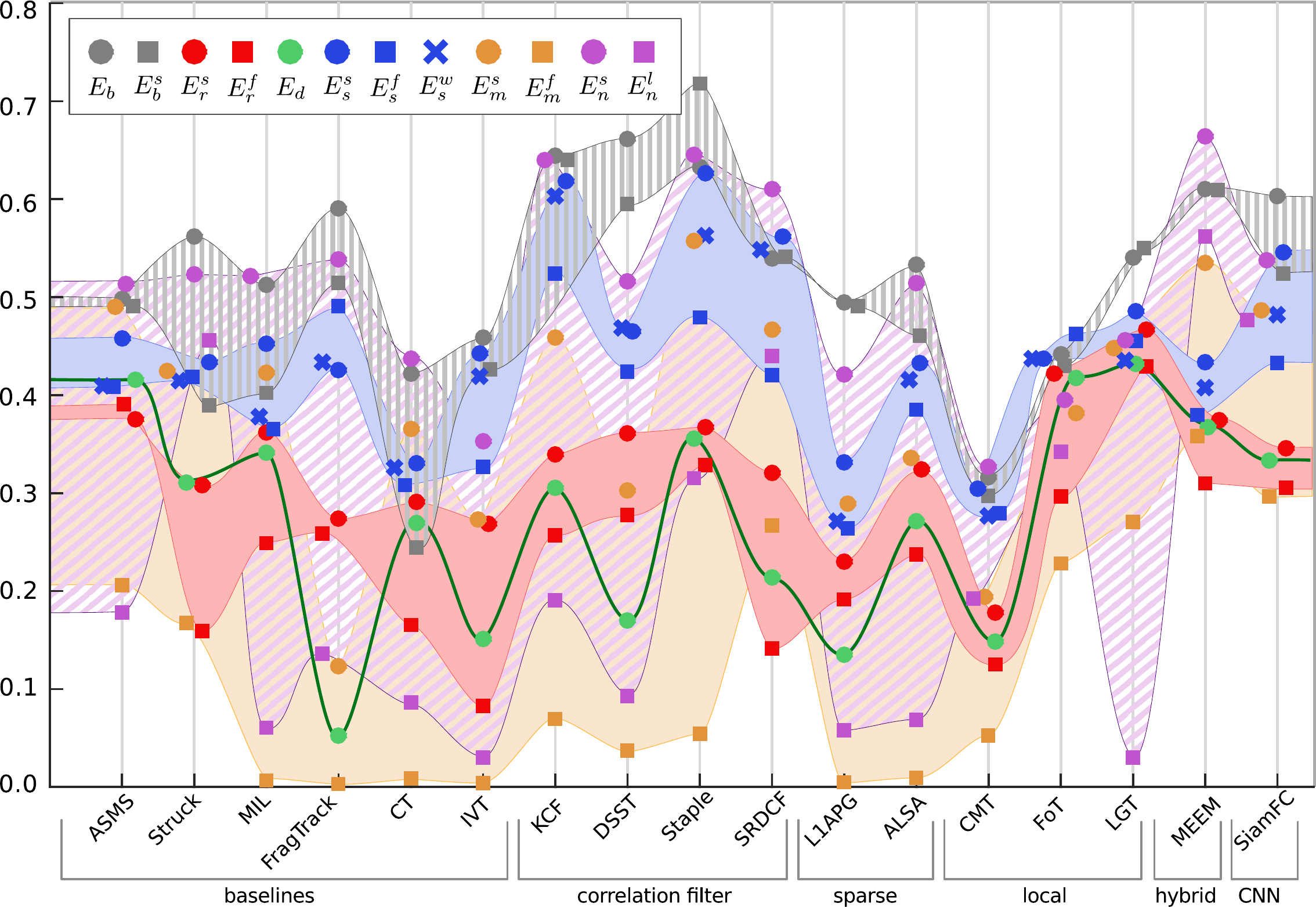}
\caption{EAO values for all motion patterns over tested trackers.}\label{fig:trackers_eao}
\end{figure}

\begin{figure}[ht]
\centering
\includegraphics[width=0.85\linewidth]{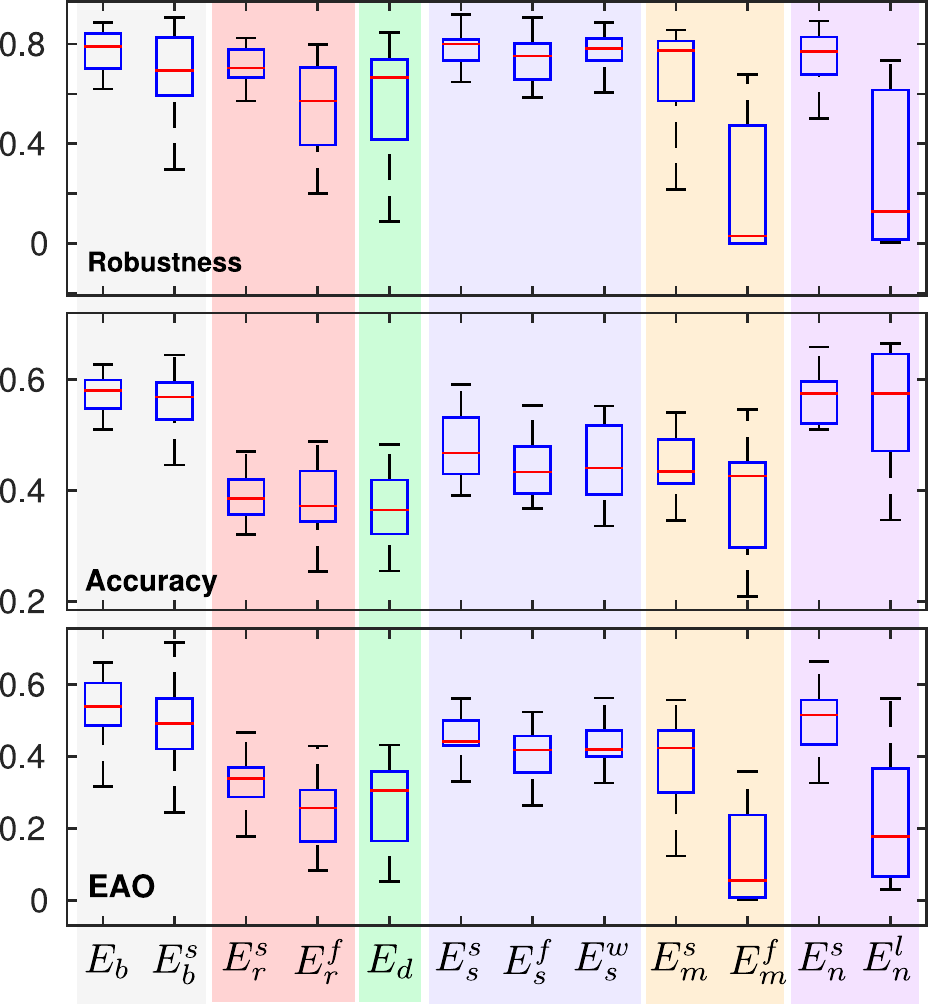}
\caption{Motion patterns difficulty levels according to robustness, accuracy, and EAO. Motion patterns are grouped by motion classes: stabilized ($E_b$), centered rotation ($E_r$), displaced rotation ($E_d$), scale change ($E_s$), planar motion ($E_m$) and noise ($E_n$).}\label{fig:types_difficulty}
\end{figure}

Results are summarized by A-R plots (Figure~\ref{fig:ar_plots}), general performance graphs (Figure~\ref{fig:trackers_eao}, Figure~\ref{fig:types_difficulty}) and in Table~\ref{tbl:eao_scores}.
In the following we discuss performance with respect to various motion pattern classes and instances.

\noindent \textbf{Scale adaptation:} Slow scale changes ($E_s^s$, Figure~\ref{fig:trackers_eao}) are addressed best by correlation filters that apply scale adaptation (i.e., KCF, DSST, Staple, SRDCF). Their performance is not significantly affected as long as the change is gradual enough, even for large amplitudes ($E_s^w$). However, fast changes ($E_s^f$) significantly reduce performance, implying that the number of scales explored should be increased in these trackers. The ConvNet tracker SiamCN does not suffer from this discrepancy, which is likely due to a large set of scales it explores. The difference in performance drop for fast ($E_s^f$) and large ($E_s^w$) scale change is low for scale-adaptive mean shift ASMS and part-based trackers (i.e., CMT, FoT and LGT). In contrast to correlation filters, these trackers do not greedily explore the scale space but apply blob size estimation (ASMS) or apply key-point-like matching approaches (CMT, FoT, LGT). Average performance at moderate scale change is better for correlation filters than part-based trackers. Struck and MEEM are least affected by scale change among the trackers that do not adapt their scale. From the AR plots in Figure~\ref{fig:ar_plots} it is apparent that the performance drops are due to a drop in accuracy, but not in failures.  

\noindent \textbf{Rotation:} Rotation ($E_r$) significantly affects performance of all tracking classes. Figure~\ref{fig:types_difficulty} and the AR plots in Figure~\ref{fig:ar_plots} show that the drop comes from a reduced accuracy as well as increased number of failures across most trackers. The drop is least apparent with ASMS, FoT and LGT which is likely due to their object visual models. The visual model in ASMS is rotation invariant since it is based on color histograms, while FoT and LGT explicitly address rotation by geometric matching. Rotation most significantly affects performance of correlation filters and ConvNets (Figure~\ref{fig:trackers_eao}). These trackers apply templates for tracking and since rotation results in significant discrepancies between the template and object, the trackers fail. In particular, from the AR plots in Figures~\ref{fig:ar_plots} we see that slow rotation ($E_r^s$) only results in decreased accuracy, but fast rotation ($E_r^f$) results in increased failures as well (e.g.,  SRDCF). 
On the other hand, the performance of correlation filters, ConvNet tracker (SiamFC) and hybrid tracker (MEEM) surpasses the part-based models when no rotation is observed ($E_b$ in Figures~\ref{fig:ar_plots} and Figure~\ref{fig:trackers_eao}). 

\noindent \textbf{Motion:} From the AR plots in Figure~\ref{fig:ar_plots} we see that slow planar motion ($E_m^s$) only slightly reduces performance in general, but this reduction is significant for most trackers in case of fast motion ($E_m^f$). LGT is the only tracker resilient to fast motion. A likely reason is the use of nearly-constant-velocity motion model in the LGT. However, the performance significantly drops for this tracker when extensive random motion is observed ($E_n^f$ in Figure~\ref{fig:trackers_eao}). Trackers like SiamFC and MEEM are least affected by all patterns of fast motions. The reason is likely in their very large search region for target localization. 
The AR plots in Figure~\ref{fig:ar_plots} indicate that SiamCF fails much more often at fast motions ($E_m^f$) than MEEM implying that MEEM is more robust at local search.

\noindent \textbf{Object size:} All trackers perform very well in the baseline setup ($E_b$) in which the object is kept centered and of constant size (Figure~\ref{fig:types_difficulty} and Figure~\ref{fig:trackers_eao}). In fact, top performance is achieved by the correlation filter trackers. The reason is that the visual model assumptions that these trackers make exactly fit this scenario. When considering smaller objects ($E_b^s$) the following trackers appear unaffected: ASMS, KCF, SRDCF, L1APG, CMT, FoT, LGT and MEEM. This implies that the level of detail of target representation in these trackers is unaffected by the reduced object size. Note that these trackers come from different classes. The AR plots in Figure~\ref{fig:ar_plots} show that performance drop in tracking small objects is most significant for baselines like CT, IVT, MIL and Fragtrac as well as a sparse tracker ASLA and the Struck tracker. The performance drop comes from increased failures, which means that their representation is not discriminative enough on this scale which leads to frequent drifts.


\noindent \textbf{General observations:} All trackers exhibit a large performance variance across the apparent-motion patterns (Figure~\ref{fig:trackers_eao}). The variance appears lowest over most motion patterns for the part-based trackers, although their average performance is moderate. Table~\ref{tbl:eao_scores} shows the average performance over the motion patterns without the baseline motion pattern ($E_b$). Among the trackers whose average EAO is within $70\%$ of best EAO are three out for four correlation filters, a hybrid tracker (MEEM), a ConvNet tracker (SiamFC), two out of three part-based trackers and two baselines (ASMS and Struck). The top three trackers in average performance are Staple ($0.470$ EOA), MEEM ($0.468$ EAO) and SiamCF ($0.448$ EAO). These trackers are also performing well on the recent benchmarks, however, our analysis shows that the weak spot of these trackers are target rotations, as well as fast movements and shaky videos.
 
\noindent \textbf{Motion class difficulty:} Considering the average EAO in Figure~\ref{fig:types_difficulty}, the most difficult classes are rotation (both patterns---central and displaced) as well as planar motion and translation noise, but the distribution of difficulty within individual classes as well as the degradation modes vary. The AR plots in Figure~\ref{fig:ar_plots} show that performance drops in rotation are due to inaccurate bounding box estimation, leading to reduced accuracy but not to complete failure. This figure also shows that trackers generally well address planar motion, but tend to fail at fast nonlinear motions due to large inter-frame displacement.

\begin{table*}[ht!]
{\footnotesize
\begin{tabular}{l|lllllllllllll}
\hline
&\textbf{ $E_b$ }&\textbf{ $E^s_b$ }&\textbf{ $E^s_r$ }&\textbf{ $E^f_r$ }&\textbf{ $E_d$ }&\textbf{ $E^{s}_s$ }&\textbf{ $E^{f}_s$ }&\textbf{ $E^{w}_s$ }&\textbf{ $E^{s}_m$ }&\textbf{ $E^{f}_m$ }&\textbf{ $E^{s}_n$ }&\textbf{ $E^{w}_n$ }&\textbf{Average}\\\hline
\TrackerSymbol{Staple} Staple & \cell{ 0.633\\\scnegs{}} & \cell{ 0.718\\\scpos{0.085}} & \cell{ 0.367\\\scnegm{-0.266}} & \cell{ 0.329\\\scnegm{-0.304}} & \cell{ 0.356\\\scnegm{-0.277}} & \cell{ 0.627\\\scnegs{-0.006}} & \cell{ 0.479\\\scnegs{-0.154}} & \cell{ 0.563\\\scnegs{-0.070}} & \cell{ 0.558\\\scnegs{-0.075}} & \cell{ 0.055\\\scnegl{-0.578}} & \cell{ 0.646\\\scpos{0.013}} & \cell{ 0.315\\\scnegl{-0.318}} & \cell{ 0.456\\\scnegm{-0.177}} \\\hline
\TrackerSymbol{MEEM} MEEM & \cell{ 0.610\\\scnegs{}} & \cell{ 0.609\\\scnegs{-0.001}} & \cell{ 0.375\\\scnegm{-0.235}} & \cell{ 0.310\\\scnegm{-0.300}} & \cell{ 0.367\\\scnegm{-0.243}} & \cell{ 0.434\\\scnegm{-0.176}} & \cell{ 0.380\\\scnegm{-0.230}} & \cell{ 0.408\\\scnegm{-0.202}} & \cell{ 0.535\\\scnegs{-0.075}} & \cell{ 0.358\\\scnegm{-0.252}} & \cell{ 0.664\\\scpos{0.054}} & \cell{ 0.562\\\scnegs{-0.048}} & \cell{ 0.455\\\scnegm{-0.155}} \\\hline
\TrackerSymbol{SiameseFC} SiamFC & \cell{ 0.603\\\scnegs{}} & \cell{ 0.525\\\scnegs{-0.078}} & \cell{ 0.346\\\scnegm{-0.257}} & \cell{ 0.305\\\scnegm{-0.298}} & \cell{ 0.334\\\scnegm{-0.269}} & \cell{ 0.546\\\scnegs{-0.057}} & \cell{ 0.433\\\scnegm{-0.170}} & \cell{ 0.482\\\scnegs{-0.120}} & \cell{ 0.487\\\scnegs{-0.116}} & \cell{ 0.296\\\scnegl{-0.307}} & \cell{ 0.538\\\scnegs{-0.065}} & \cell{ 0.477\\\scnegs{-0.126}} & \cell{ 0.433\\\scnegm{-0.169}} \\\hline
\TrackerSymbol{KCF} KCF & \cell{ 0.644\\\scnegs{}} & \cell{ 0.640\\\scnegs{-0.004}} & \cell{ 0.339\\\scnegm{-0.305}} & \cell{ 0.257\\\scnegl{-0.387}} & \cell{ 0.306\\\scnegl{-0.338}} & \cell{ 0.618\\\scnegs{-0.026}} & \cell{ 0.524\\\scnegs{-0.120}} & \cell{ 0.603\\\scnegs{-0.041}} & \cell{ 0.459\\\scnegm{-0.185}} & \cell{ 0.069\\\scnegl{-0.575}} & \cell{ 0.640\\\scnegs{-0.005}} & \cell{ 0.191\\\scnegl{-0.453}} & \cell{ 0.422\\\scnegm{-0.222}} \\\hline
\TrackerSymbol{SRDCF} SRDCF & \cell{ 0.539\\\scnegs{}} & \cell{ 0.542\\\scpos{0.002}} & \cell{ 0.320\\\scnegm{-0.219}} & \cell{ 0.141\\\scnegl{-0.398}} & \cell{ 0.214\\\scnegl{-0.326}} & \cell{ 0.562\\\scpos{0.022}} & \cell{ 0.420\\\scnegs{-0.119}} & \cell{ 0.548\\\scpos{0.009}} & \cell{ 0.467\\\scnegs{-0.073}} & \cell{ 0.267\\\scnegl{-0.272}} & \cell{ 0.610\\\scpos{0.071}} & \cell{ 0.440\\\scnegs{-0.099}} & \cell{ 0.412\\\scnegs{-0.128}} \\\hline
\TrackerSymbol{LGT} LGT & \cell{ 0.540\\\scnegs{}} & \cell{ 0.550\\\scpos{0.010}} & \cell{ 0.467\\\scnegs{-0.074}} & \cell{ 0.429\\\scnegs{-0.111}} & \cell{ 0.432\\\scnegs{-0.109}} & \cell{ 0.486\\\scnegs{-0.055}} & \cell{ 0.455\\\scnegs{-0.085}} & \cell{ 0.435\\\scnegs{-0.105}} & \cell{ 0.448\\\scnegs{-0.092}} & \cell{ 0.271\\\scnegm{-0.270}} & \cell{ 0.456\\\scnegs{-0.085}} & \cell{ 0.030\\\scnegl{-0.510}} & \cell{ 0.405\\\scnegm{-0.135}} \\\hline
\TrackerSymbol{ASMS} ASMS & \cell{ 0.498\\\scnegs{}} & \cell{ 0.491\\\scnegs{-0.007}} & \cell{ 0.375\\\scnegs{-0.123}} & \cell{ 0.390\\\scnegs{-0.107}} & \cell{ 0.415\\\scnegs{-0.082}} & \cell{ 0.458\\\scnegs{-0.040}} & \cell{ 0.409\\\scnegs{-0.089}} & \cell{ 0.410\\\scnegs{-0.088}} & \cell{ 0.490\\\scnegs{-0.008}} & \cell{ 0.206\\\scnegl{-0.292}} & \cell{ 0.514\\\scpos{0.016}} & \cell{ 0.178\\\scnegl{-0.320}} & \cell{ 0.394\\\scnegs{-0.104}} \\\hline
\TrackerSymbol{FoT} FoT & \cell{ 0.442\\\scnegs{}} & \cell{ 0.430\\\scnegs{-0.011}} & \cell{ 0.422\\\scnegs{-0.020}} & \cell{ 0.297\\\scnegm{-0.145}} & \cell{ 0.417\\\scnegs{-0.024}} & \cell{ 0.437\\\scnegs{-0.005}} & \cell{ 0.462\\\scpos{0.021}} & \cell{ 0.437\\\scnegs{-0.004}} & \cell{ 0.381\\\scnegs{-0.060}} & \cell{ 0.228\\\scnegm{-0.213}} & \cell{ 0.395\\\scnegs{-0.046}} & \cell{ 0.342\\\scnegs{-0.100}} & \cell{ 0.386\\\scnegs{-0.055}} \\\hline
\TrackerSymbol{Struck} Struck & \cell{ 0.562\\\scnegs{}} & \cell{ 0.390\\\scnegm{-0.172}} & \cell{ 0.308\\\scnegm{-0.254}} & \cell{ 0.160\\\scnegl{-0.403}} & \cell{ 0.311\\\scnegm{-0.251}} & \cell{ 0.433\\\scnegs{-0.129}} & \cell{ 0.418\\\scnegm{-0.144}} & \cell{ 0.414\\\scnegm{-0.148}} & \cell{ 0.425\\\scnegs{-0.137}} & \cell{ 0.167\\\scnegl{-0.395}} & \cell{ 0.524\\\scnegs{-0.038}} & \cell{ 0.456\\\scnegs{-0.106}} & \cell{ 0.364\\\scnegm{-0.198}} \\\hline
\TrackerSymbol{DSST} DSST & \cell{ 0.662\\\scnegs{}} & \cell{ 0.595\\\scnegs{-0.066}} & \cell{ 0.361\\\scnegm{-0.301}} & \cell{ 0.278\\\scnegl{-0.384}} & \cell{ 0.170\\\scnegl{-0.492}} & \cell{ 0.465\\\scnegm{-0.197}} & \cell{ 0.423\\\scnegm{-0.238}} & \cell{ 0.468\\\scnegm{-0.193}} & \cell{ 0.303\\\scnegl{-0.359}} & \cell{ 0.038\\\scnegl{-0.624}} & \cell{ 0.516\\\scnegs{-0.146}} & \cell{ 0.093\\\scnegl{-0.568}} & \cell{ 0.337\\\scnegm{-0.324}} \\\hline
\TrackerSymbol{MIL} MIL & \cell{ 0.513\\\scnegs{}} & \cell{ 0.402\\\scnegs{-0.110}} & \cell{ 0.362\\\scnegm{-0.151}} & \cell{ 0.249\\\scnegl{-0.264}} & \cell{ 0.341\\\scnegm{-0.172}} & \cell{ 0.452\\\scnegs{-0.060}} & \cell{ 0.366\\\scnegm{-0.147}} & \cell{ 0.378\\\scnegm{-0.135}} & \cell{ 0.423\\\scnegs{-0.089}} & \cell{ 0.006\\\scnegl{-0.506}} & \cell{ 0.522\\\scpos{0.009}} & \cell{ 0.061\\\scnegl{-0.452}} & \cell{ 0.324\\\scnegm{-0.189}} \\\hline
\TrackerSymbol{ASLA} ASLA & \cell{ 0.533\\\scnegs{}} & \cell{ 0.461\\\scnegs{-0.072}} & \cell{ 0.324\\\scnegm{-0.209}} & \cell{ 0.237\\\scnegl{-0.296}} & \cell{ 0.271\\\scnegm{-0.262}} & \cell{ 0.433\\\scnegs{-0.100}} & \cell{ 0.385\\\scnegm{-0.148}} & \cell{ 0.416\\\scnegs{-0.117}} & \cell{ 0.336\\\scnegm{-0.196}} & \cell{ 0.009\\\scnegl{-0.524}} & \cell{ 0.514\\\scnegs{-0.019}} & \cell{ 0.069\\\scnegl{-0.464}} & \cell{ 0.314\\\scnegm{-0.219}} \\\hline
\TrackerSymbol{FragTrack} FragTrack & \cell{ 0.590\\\scnegs{}} & \cell{ 0.514\\\scnegs{-0.076}} & \cell{ 0.274\\\scnegl{-0.316}} & \cell{ 0.259\\\scnegl{-0.331}} & \cell{ 0.052\\\scnegl{-0.538}} & \cell{ 0.426\\\scnegm{-0.164}} & \cell{ 0.491\\\scnegs{-0.099}} & \cell{ 0.434\\\scnegm{-0.157}} & \cell{ 0.123\\\scnegl{-0.467}} & \cell{ 0.003\\\scnegl{-0.587}} & \cell{ 0.539\\\scnegs{-0.052}} & \cell{ 0.136\\\scnegl{-0.454}} & \cell{ 0.296\\\scnegm{-0.295}} \\\hline
\TrackerSymbol{CT} CT & \cell{ 0.422\\\scnegs{}} & \cell{ 0.245\\\scnegm{-0.177}} & \cell{ 0.291\\\scnegm{-0.131}} & \cell{ 0.166\\\scnegl{-0.256}} & \cell{ 0.269\\\scnegm{-0.153}} & \cell{ 0.331\\\scnegs{-0.091}} & \cell{ 0.308\\\scnegm{-0.114}} & \cell{ 0.326\\\scnegs{-0.096}} & \cell{ 0.366\\\scnegs{-0.056}} & \cell{ 0.008\\\scnegl{-0.413}} & \cell{ 0.437\\\scpos{0.015}} & \cell{ 0.086\\\scnegl{-0.335}} & \cell{ 0.258\\\scnegm{-0.164}} \\\hline
\TrackerSymbol{IVT} IVT & \cell{ 0.458\\\scnegs{}} & \cell{ 0.427\\\scnegs{-0.032}} & \cell{ 0.269\\\scnegm{-0.190}} & \cell{ 0.083\\\scnegl{-0.376}} & \cell{ 0.151\\\scnegl{-0.308}} & \cell{ 0.442\\\scnegs{-0.016}} & \cell{ 0.327\\\scnegm{-0.131}} & \cell{ 0.419\\\scnegs{-0.039}} & \cell{ 0.273\\\scnegm{-0.186}} & \cell{ 0.004\\\scnegl{-0.455}} & \cell{ 0.353\\\scnegs{-0.106}} & \cell{ 0.030\\\scnegl{-0.428}} & \cell{ 0.252\\\scnegm{-0.206}} \\\hline
\TrackerSymbol{L1APG} L1APG & \cell{ 0.495\\\scnegs{}} & \cell{ 0.491\\\scnegs{-0.004}} & \cell{ 0.230\\\scnegl{-0.265}} & \cell{ 0.192\\\scnegl{-0.303}} & \cell{ 0.135\\\scnegl{-0.360}} & \cell{ 0.332\\\scnegm{-0.163}} & \cell{ 0.264\\\scnegm{-0.231}} & \cell{ 0.272\\\scnegm{-0.223}} & \cell{ 0.289\\\scnegm{-0.206}} & \cell{ 0.005\\\scnegl{-0.490}} & \cell{ 0.421\\\scnegs{-0.074}} & \cell{ 0.058\\\scnegl{-0.437}} & \cell{ 0.244\\\scnegl{-0.251}} \\\hline
\TrackerSymbol{CMT} CMT & \cell{ 0.316\\\scnegs{}} & \cell{ 0.297\\\scnegs{-0.019}} & \cell{ 0.178\\\scnegm{-0.139}} & \cell{ 0.125\\\scnegl{-0.191}} & \cell{ 0.148\\\scnegl{-0.168}} & \cell{ 0.304\\\scnegs{-0.012}} & \cell{ 0.280\\\scnegs{-0.036}} & \cell{ 0.277\\\scnegs{-0.039}} & \cell{ 0.194\\\scnegm{-0.122}} & \cell{ 0.053\\\scnegl{-0.264}} & \cell{ 0.327\\\scpos{0.011}} & \cell{ 0.192\\\scnegm{-0.124}} & \cell{ 0.216\\\scnegm{-0.100}} \\\hline
Average & \cell{ 0.533\\\scnegs{}} & \cell{ 0.490\\\scnegs{-0.043}} & \cell{ 0.330\\\scnegm{-0.203}} & \cell{ 0.247\\\scnegl{-0.286}} & \cell{ 0.276\\\scnegm{-0.257}} & \cell{ 0.458\\\scnegs{-0.075}} & \cell{ 0.401\\\scnegs{-0.131}} & \cell{ 0.429\\\scnegs{-0.104}} & \cell{ 0.386\\\scnegm{-0.147}} & \cell{ 0.120\\\scnegl{-0.413}} & \cell{ 0.507\\\scnegs{-0.026}} & \cell{ 0.219\\\scnegl{-0.314}} & \cell{ \\\scnegs{}} \\\hline
\end{tabular}
}
\caption{Overview of the EAO scores and their relative differences according to the baseline score. The top value in each cell represents the absolute EAO score while the bottom one represents the EAO difference in relation to the baseline experiment. Green text denotes relative increase, orange text relative decrease, and red and bold red text decrease greater than $25\%$ and $50\%$ of the baseline score. The baseline experiment is not used forfor computing the average tracker score.}\label{tbl:eao_scores}
\end{table*}

\subsection{Relation to existing benchmarks}

A relation of AMP to the existing tracking benchmarks was established by comparing the ranks of common trackers on a well known OTB100~\cite{otb_pami2015} and the recent UAV123~\cite{uav_benchmark_eccv2016} benchmark.

\begin{figure}[ht]
\centering
\includegraphics[width=\linewidth]{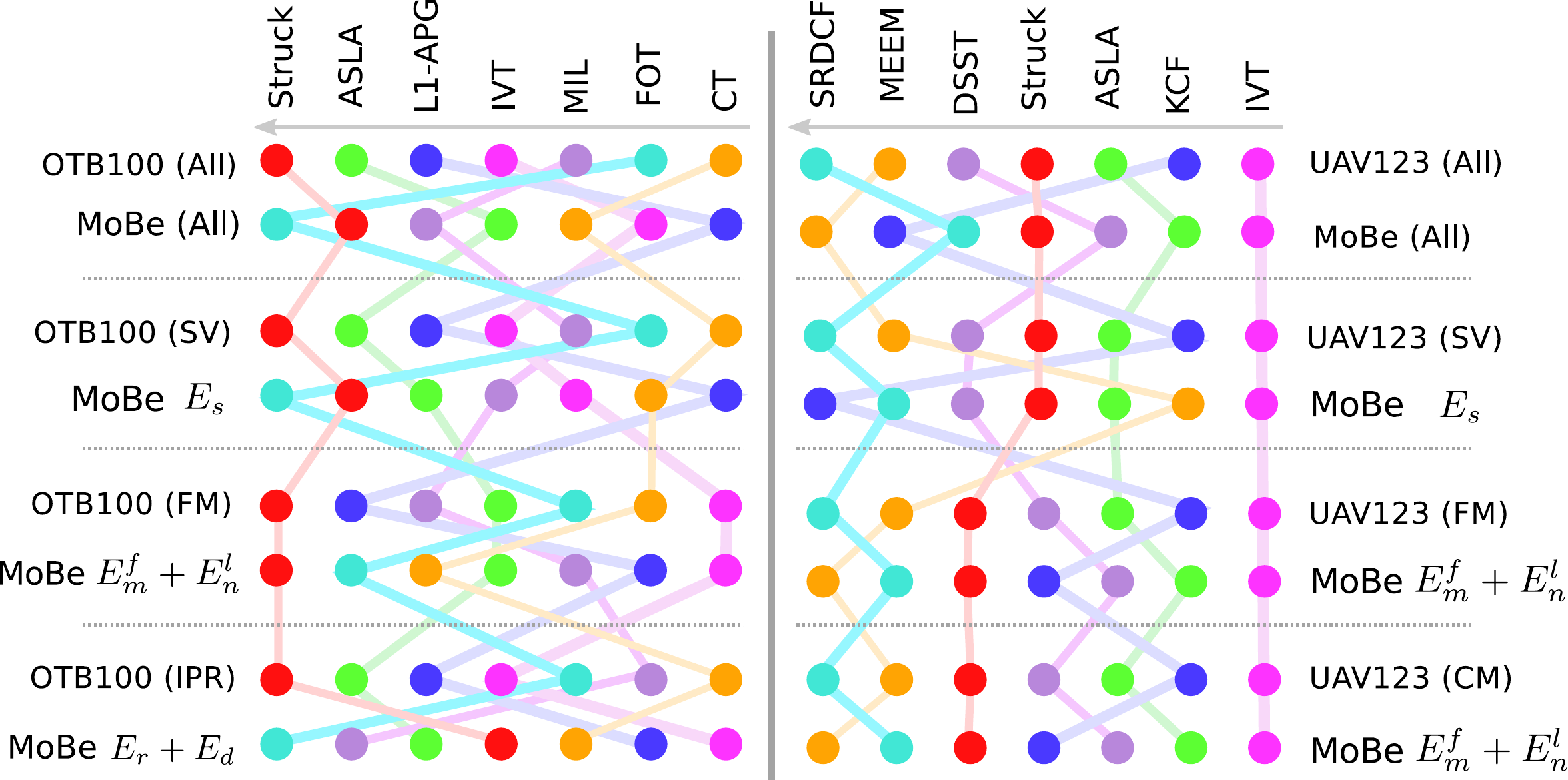}
\caption{Tracker ranking comparison of AMP with OTB100 (left) and UAV123 (right). The trackers are sorted from left (best) to right. Attribute abbreviations: SV -- Scale Variation, FM -- Fast Motion, IPT -- In-Plane Rotation, CM -- Camera Motion.}\label{fig:benchmark_comparison}
\end{figure}

 \textbf{Comparison with OTB100:} The OTB100 contains relatively old trackers, therefore the intersection is in the following six trackers: ASLA, CT, FoT, IVT, L1-APG, MIL, and Struck. Figure~\ref{fig:benchmark_comparison} shows the ranking differences between these trackers for the different ranking modes. The ranking by average performance differs mainly L1-APG and FoT trackers. The possible reasons for this are different implementations, algorithm parameters, as well as different evaluation methodology\footnote{The OTB100 methodology does not restart tracker on failure which can lead to large differences between trackers that support re-detection and those that do not.}. Three motion patterns in OTB100 are compatible with MAP: scale variation, fast motion and in-plane rotation. Performance over three scale changing motion patterns on AMP was averaged to obtain a scale change ranking. While the FoT achieves top performance on AMP it is positioned relatively low on OTB100, which is likely due to different implementations (ours is from the authors) and interaction of other attributes on OTB100. Both rankings place Struck at the top, ranks of other trackers vary. The fast motion ranking on AMP was obtained by averaging fast motions, i.e., $E_m^f$ and $E_n^w$. Both benchmarks rank Struck as top performing and IVT as worst performing. The in-plane rotation attribute was compared with combined ranking of center and displaced rotation ($E_r$ and $E_d$). The situation is similar to scale change, where FoT, which explicitly addresses rotation is ranked much lower according to OTB100.

 \textbf{Comparison with UAV123:} The AMP and UAV123 intersect in the following seven trackers: ASLA, DSST, IVT, KCF, MEEM, SRDCF, and Struck. The comparison of average performance as well as with respect to three types of motion patterns: scale variation, camera motion and fast motion is shown in
Figure~\ref{fig:benchmark_comparison}. The ranks are mostly consistent, the best two trackers are mostly SRDCF and MEEM. 
A discrepancy is observed for the MEEM tracker at scale change attribute. MEEM does not adapt the scale, which results in low rank at AMP. But is ranked high on UAV123, which is likely due to attribute cross-talk. The discrepancy in KCF is due to implementation -- our KCF adapts scale. Notice that the UAV123 ranks on camera motion are equal to scale variation ranks. We therefore compare both with ranks obtained by averaging fast in-plane motion ($E_m^f$) and large translation noise ($E_n^w$) performance on AMP. The ranks match very well, which means that AMP offers a significant level of granularity in analysis.

\section{Discussion and conclusions}
\label{sec:conclution} 

We have proposed a novel approach for single-target tracker evaluation on parameterized motion-related attributes. At the core of our approach is the use of 360 degree videos to generate annotated realistic-looking tracking scenarios. We have presented a novel benchmark AMP, composed of annotated dataset of fifteen such videos and the results of 17 state-of-the-art trackers. We have experimentally verified the realism of the generated sequences by reproducing partial ranks available in standard benchmarks.

The results of our experiments provide a detailed overview of strengths and limitations of modern short-term visual trackers. The scale change appears to be well addressed by many tracking approaches. Even trackers that do not adapt scale do not fail often. Nevertheless, in practice scale change is often accompanied by appearance change or fast motion, which increase chances of failures. We believe that this is the reason why scale change is perceived as a challenging attribute in related benchmarks. The state-of-the-art trackers perform reasonably well in tracking small targets. Rotation and abrupt motion are two of the most challenging motion classes. Due to their scarcity on existing benchmarks they remain poorly addressed by most modern trackers. Our results have shown that non-random motions are well addressed by motion models, which have also become rare in modern trackers. We believe that future research in tracker development should focus on these topics to make further improvements.

We have demonstrated the usefulness of the proposed approach for evaluating trackers in a controlled, yet realistic environment. The approach is complementary to existing benchmarks allowing better insights into tracking behavior on various apparent-motion patterns. Moreover, capturing omnidirectional videos is nowadays possible with commodity equipment. Therefore our dataset adaptation to a specific tracking scenario may in fact be easier than in traditional approaches since it does not require careful planning before the acquisition to cover all possible motion patterns.

Our future work will focus on generalizing our framework to complex motion patterns and their effects on tracking performance. We also plan to explore adaptation of our evaluation methodology to active tracking.


{\small
\bibliographystyle{ieee}
\bibliography{bib}
}

\end{document}